\documentclass{article}
\usepackage[final]{neurips_2023,listings}
\usepackage[utf8]{inputenc} % Ensure correct encoding

\title{Cognitive Prompts Using Guilford’s Structure of Intellect Model}

\author{
    Oliver Kramer \\
    CI Lab, Department of Computer Science\\University of Oldenburg, Germany \\
    \texttt{oliver.kramer@uni-oldenburg.de}
}

\begin{document}
\maketitle

\begin{abstract}
Large language models (LLMs) demonstrate strong language generation capabilities but often struggle with structured reasoning, leading to inconsistent or suboptimal problem-solving. To mitigate this limitation, Guilford’s Structure of Intellect (SOI) model—a foundational framework from intelligence theory—is leveraged as the basis for cognitive prompt engineering. The SOI model categorizes cognitive operations such as pattern recognition, memory retrieval, and evaluation, offering a systematic approach to enhancing LLM reasoning and decision-making. This position paper presents a novel cognitive prompting approach for enforcing SOI-inspired reasoning for improving clarity, coherence, and adaptability in model responses.
\end{abstract}

\section{Introduction}

Large language models (LLMs) have demonstrated remarkable capabilities in generating and understanding text, yet they often struggle with structured reasoning and systematic problem-solving. While these models excel at language-based tasks, they frequently produce inconsistent or suboptimal solutions when faced with complex problems requiring logical progression and structured analysis. One of the key challenges in LLM reasoning is the lack of explicit cognitive strategies, leading to responses that may lack coherence, adaptability, or depth. Addressing this issue requires a systematic approach that enables models to employ structured cognitive operations akin to human problem-solving strategies.

To this end, a novel prompting approach is introduced that explicitly guides LLMs using Guilford’s Structure of Intellect (SOI) model. The SOI model has evolved over time, with key contributions from Guilford \cite{guilford1,guilford2,guilford3}. The framework categorizes cognitive abilities into three key dimensions: operations, contents, and products. Operations define cognitive functions such as pattern recognition, memory retrieval, divergent ideation, convergent reasoning, and evaluation. Contents represent the type of information being processed, while products describe the structure of the output generated. By incorporating this model into prompt engineering, LLMs can systematically select cognitive abilities and apply them dynamically in problem-solving scenarios.

\section{Guilford's Structure of Intellect Model}

Guilford's SOI model presents a comprehensive framework
for understanding cognitive abilities by categorizing human intelligence into three dimensions: operations, contents, and products. 

%The operations dimension describes five fundamental cognitive processes: cognition, memory, divergent production, convergent production, and evaluation.

The operations dimension in Guilford's SOI model defines five fundamental cognitive processes:  

\begin{itemize}
    \item \textbf{Cognition}: The ability to perceive, recognize, and comprehend information. It includes identifying patterns, understanding relationships, and interpreting meanings within various contexts. Cognition is essential for acquiring knowledge and serves as the foundation for higher-order thinking.
    
    \item \textbf{Memory}: The process of encoding, storing, and retrieving information. It encompasses both short-term and long-term memory functions, enabling individuals to recall facts, concepts, and experiences when needed. Effective memory functions contribute to learning, problem-solving, and decision-making.
    
    \item \textbf{Divergent Production}: The ability to generate multiple, diverse responses to a given problem or stimulus. It involves creative thinking, brainstorming, and exploring various possibilities without immediate judgment. This cognitive process is fundamental in innovation, artistic expression, and scientific discovery.
    
    \item \textbf{Convergent Production}: The process of narrowing down multiple possibilities to find the single best or most logical solution to a problem. It requires critical thinking, reasoning, and systematic evaluation to determine the most effective answer. This type of thinking is crucial for problem-solving in structured and rule-based domains.
    
    \item \textbf{Evaluation}: The capacity to assess the accuracy, validity, and effectiveness of information, ideas, or solutions. It involves making judgments based on criteria, evidence, or logical consistency. Evaluation plays a key role in decision-making, self-regulation, and critical analysis.
\end{itemize}

The contents dimension in Guilford's SOI model categorizes the types of information being processed into five distinct forms:

\begin{itemize}
    \item \textbf{Figural-Visual}: Information that is represented in the form of static images, shapes, diagrams, or spatial patterns. This includes the recognition and manipulation of visual elements such as pictures, graphs, and geometric figures. Figural-visual processing is crucial in fields such as architecture, engineering, and visual arts.
    
    \item \textbf{Figural-Auditory}: Information that is perceived and processed in the form of sounds, tones, rhythms, or spoken words. This category encompasses auditory pattern recognition, music perception, and phonological processing. It plays a fundamental role in language comprehension, musical ability, and auditory memory.
    
    \item \textbf{Symbolic}: Information that is represented through abstract symbols, numbers, codes, or notations. This includes mathematical expressions, computer programming languages, and logic-based symbols. Symbolic processing is essential for disciplines such as mathematics, cryptography, and formal logic.
    
    \item \textbf{Semantic}: Information that is expressed through words, concepts, and meaningful linguistic structures. This involves understanding definitions, reading comprehension, and verbal reasoning. Semantic processing is foundational to communication, literature, and knowledge acquisition.
    
    \item \textbf{Behavioral}: Information that pertains to human interactions, emotions, and social cues. It involves the recognition and interpretation of nonverbal gestures, facial expressions, body language, and interpersonal dynamics. Behavioral content is critical in psychology, sociology, leadership, and social intelligence.
\end{itemize}

The products dimension in Guilford's SOI model describes the different forms in which cognitive processing generates outcomes. These are classified into six distinct categories:

\begin{itemize}
    \item \textbf{Units}: The smallest elements of information that are processed as discrete, singular entities. Examples include individual letters, numbers, words, sounds, or objects. Recognizing and recalling units is fundamental to basic perception, memory, and knowledge acquisition.

    \item \textbf{Classes}: Groups or categories that organize multiple units based on shared characteristics or common properties. This includes recognizing patterns, identifying conceptual similarities, and classifying objects, words, or ideas into meaningful sets. Class formation is crucial for learning, taxonomy, and generalization.

    \item \textbf{Relations}: The connections between two or more units or classes based on identifiable associations. These can include cause-effect relationships, analogies, hierarchies, sequences, or logical linkages. The ability to discern relations is essential for reasoning, problem-solving, and inferential thinking.

    \item \textbf{Systems}: Complex structures composed of multiple interrelated units, classes, and relations. Systems involve the organization of elements into an integrated whole, such as frameworks, networks, or conceptual models. Understanding systems is key in disciplines such as engineering, organizational theory, and strategic planning.

    \item \textbf{Transformations}: The ability to modify, alter, or reinterpret existing information to create new versions or perspectives. This includes linguistic paraphrasing, reconfiguring equations, reimagining artistic designs, and adapting ideas for novel applications. Transformations are central to creativity, innovation, and problem-solving.

    \item \textbf{Implications}: The predictions, inferences, or conclusions drawn from a given set of information. Implications involve extending known information to anticipate outcomes, deduce consequences, or propose theoretical possibilities. This form of thinking is fundamental to decision-making, scientific hypothesis generation, and strategic forecasting.
\end{itemize}

By structuring intelligence along these three axes, Guilford’s model provides a systematic approach to analyzing and enhancing cognitive processes in problem-solving and creative thinking.

\section{Cognitive Prompt Strategy}

A prompt strategy based on Guilford’s SOI model is introduced to guide structured reasoning in large language models. This approach systematically decomposes problem-solving into three dimensions: cognitive operations, types of information, and products. By structuring responses according to these categories, language models can follow explicit cognitive steps to enhance clarity, coherence, and adaptability.

\subsection{SOI Prompt}

The following prompt instructs the cognitive problem-solving agent to analyze and solve complex problems using the SOI framework:

\begin{lstlisting}
You are a cognitive problem-solving agent utilizing Guilford's Structure of Intellect (SOI) model to analyze and solve complex problems systematically. Your reasoning follows structured cognitive operations across three key dimensions: operations, types of information, and products.

When presented with a problem, structure your response using the SOI framework:

1. Cognitive Operation (Select One)
   - Cognition: Identify key patterns and relationships in the problem.
   - Memory: Recall relevant economic, technological, and policy concepts.
   - Divergent Production: Generate multiple solution strategies.
   - Convergent Production: Identify the single best course of action.
   - Evaluation: Assess potential solutions based on effectiveness and feasibility.

2. Type of Information (Select One)
   - Figural-Visual: Use diagrams or spatial representations if applicable.
   - Figural-Auditory: Consider spoken, tonal, or rhythm-based aspects.
   - Symbolic: Use mathematical, computational, or coded representations.
   - Semantic: Interpret linguistic and conceptual meanings.
   - Behavioral: Analyze human interactions, policies, and socio-economic impacts.

3. Product (Select One)
   - Units: Identify fundamental concepts within the problem.
   - Classes: Categorize the economic trends or policy challenges.
   - Relations: Define cause-effect relationships and interactions.
   - Systems: Construct a framework integrating economic, technological, and policy factors.
   - Transformations: Suggest modifications or adaptations to address any emerging challenges.
   - Implications: Predict long-term effects and propose strategic recommendations.
\end{lstlisting}

\pagebreak

\subsection{Illustrative Example: Step-by-Step SOI Analysis}

To illustrate this framework, consider the following problem using GPT-4o by OpenAI.

\textbf{Problem Statement:}

\begin{quote}
“In recent years, the global economy has experienced significant shifts due to rapid technological advances, increased globalization, and evolving consumer behaviors. Emerging markets have become important drivers of growth, while traditional industries face disruptions from innovative startups. Policy makers are challenged to balance economic growth with social and environmental sustainability.”
\end{quote}

\noindent Using the SOI model, the problem is analyzed step by step:

\textbf{Step 1: Select Cognitive Operation}

\begin{lstlisting}
Cognitive Operation: Cognition
Rationale: The goal is to identify key patterns and relationships between technological advancements, globalization, and economic shifts.
\end{lstlisting}

\textbf{Step 2: Select Type of Information}

\begin{lstlisting}
Type of Information: Semantic
Rationale: The problem involves interpreting linguistic and conceptual meanings, requiring an understanding of economic and policy-related trends.
\end{lstlisting}

\textbf{Step 3: Select Product}

\begin{lstlisting}
Product: Relations
Rationale: The problem requires defining cause-effect relationships between technological changes, market dynamics, and policy-making challenges.
\end{lstlisting}

\textbf{Step 4: Generate a Structured Response}

\begin{lstlisting}
Technological advancements and globalization have created a dynamic interplay between emerging markets and traditional industries. Innovations drive efficiency and new consumer behaviors, enabling startups to challenge legacy sectors. As a result, traditional industries must adapt or risk obsolescence. The rise of emerging economies contributes to economic diversification but also increases competitive pressures. Policy makers face the challenge of balancing economic expansion with sustainability goals, necessitating regulatory frameworks that foster both growth and environmental responsibility. These relationships illustrate the interconnected nature of economic, technological, and policy-driven transformations.
\end{lstlisting}

This example demonstrates how structured prompts using SOI principles enhance the reasoning process in LLMs. By explicitly defining cognitive operations, information types, and expected outputs, the model follows a structured pathway, improving the clarity and interpretability of its responses.

\section{Related Work}
\label{sec:rel}

Recent advances in prompt engineering have underscored the importance of guiding LLMs through structured reasoning processes. Chain-of-Thought (CoT) prompting \cite{wei2022chain} pioneered this area by decomposing complex problems into explicit, step-by-step reasoning sequences, which helps LLMs produce more coherent and interpretable outputs. Building on this foundation, cognitive prompting \cite{kramer2025cognitive} structures prompts around distinct cognitive operations, thereby aligning model reasoning with human-like strategies and enhancing generalization across diverse tasks. The ReAct framework \cite{yao2022react} further integrates reasoning with real-time decision-making, enabling dynamic adaptation as the problem context evolves.

Complementing these strategies, optimization-driven prompting techniques have emerged to refine LLM performance further. For example, Prompt Breeder \cite{promptbreeder2022} leverages evolutionary algorithms to iteratively evolve and optimize prompt formulations, while Automated Prompt Engineering (APE) \cite{ape} and Optimization by PROmpting (OPRO) \cite{opro} systematically apply optimization algorithms to fine-tune instructions. Collectively, these approaches demonstrate that structured and adaptive prompt engineering significantly enhances the problem-solving capabilities and efficiency of LLMs.

\section{Conclusion}

This paper introduces a novel cognitive prompting framework for LLMs, explicitly integrating Guilford’s SOI model into prompt engineering. By structuring problem-solving around the dimensions of operations, contents, and products, a systematic approach is developed that enables LLMs to dynamically select and apply cognitive abilities. This method ensures structured reasoning, guiding models through problem analysis, information retrieval, creative ideation, and decision-making in a way that mirrors human cognitive strategies. 
This work establishes a systematic and theoretically grounded foundation for cognitive prompt engineering, bridging classical psychological models with modern AI. By formalizing structured reasoning processes in LLMs, this approach supports the advancement of interpretable, adaptive, and cognitively aligned artificial intelligence.

Future work can explore the refinement of cognitive prompting by incorporating reinforcement learning techniques to optimize the selection of cognitive operations dynamically. Expanding the framework to support multimodal reasoning, integrating visual and auditory processing alongside textual prompts, could further enhance adaptability in complex problem-solving scenarios. Additionally, evaluating SOI-based prompting across a broader range of real-world applications, such as scientific discovery, legal reasoning, and policy analysis, would provide deeper insights into its effectiveness and scalability.

\bibliographystyle{plainnat}  % Standard NeurIPS-compatible style

\begin{thebibliography}{1}

\bibitem{guilford1}
J.~P. Guilford.
\newblock The structure of intellect.
\newblock {\em Psychological Bulletin}, 53(4):267--293, 1956.

\bibitem{guilford2}
J.~P. Guilford.
\newblock {\em The Nature of Human Intelligence}.
\newblock McGraw-Hill, 1967.

\bibitem{guilford3}
J.~P. Guilford.
\newblock Some changes in the structure of intellect model.
\newblock {\em Educational and Psychological Measurement}, 48(1):1--4, 1988.

\bibitem{wei2022chain}
Jason Wei, Xuezhi Wang, Dale Schuurmans, Maarten Bosma, Brian Ichter, Fei Xia,
  Ed~H. Chi, Quoc~V. Le, and Denny Zhou.
\newblock Chain-of-thought prompting elicits reasoning in large language
  models.
\newblock In Sanmi Koyejo, S.~Mohamed, A.~Agarwal, Danielle Belgrave, K.~Cho,
  and A.~Oh, editors, {\em Neural Information Processing Systems (NeurIPS)
  Workshop}, volume~35, pages 24824--24837, 2022.

\bibitem{kramer2025cognitive}
Oliver Kramer and Jill Baumann.
\newblock Unlocking structured thinking in language models with cognitive
  prompting.
\newblock In {\em European Symposium on Artificial Neural Network (ESANN)},
  2025.

\bibitem{yao2022react}
Shunyu Yao, Jeffrey Zhao, Dian Yu, Nan Du, Izhak Shafran, Karthik~R.
  Narasimhan, and Yuan Cao.
\newblock React: Synergizing reasoning and acting in language models.
\newblock In {\em International Conference on Learning Representations (ICLR)},
  2023.

\bibitem{promptbreeder2022}
Chrisantha Fernando, Dylan Banarse, Henryk Michalewski, Simon Osindero, and Tim
  Rocktäschel.
\newblock Promptbreeder: Self-referential self-improvement via prompt
  evolution.
\newblock {\em Neural Information Processing Systems (NeurIPS) Workshop}, 2023.

\bibitem{ape}
Yongchao Zhou, Andrei~Ioan Muresanu, Ziwen Han, Keiran Paster, Silviu Pitis,
  Harris Chan, and Jimmy Ba.
\newblock Large language models are human-level prompt engineers.
\newblock In {\em International Conference on Learning Representations (ICLR)},
  2023.

\bibitem{opro}
Chengrun Yang, Xuezhi Wang, Yifeng Lu, Hanxiao Liu, Quoc~V Le, Denny Zhou, and
  Xinyun Chen.
\newblock Large language models as optimizers.
\newblock In {\em International Conference on Learning Representations (ICLR)},
  2024.

\end{thebibliography}

\end{document}